# Progressive Sampling-Based Bayesian Optimization for Efficient and Automatic Machine Learning Model Selection

**Xueqiang Zeng**[a], **Gang Luo**[b]

[a]Computer Center, Nanchang University, 999 Xuefu Road, Nanchang, Jiangxi Province 330031, P.R. China

xqzeng@ncu.edu.cn

[b]Department of Biomedical Informatics and Medical Education, University of Washington, UW Medicine South Lake Union, 850 Republican Street, Building C, Box 358047, Seattle, WA 98109, USA

luogang@uw.edu

**Corresponding author**:

Gang Luo, PhD (ORCID: 0000-0001-7217-4008)

Department of Biomedical Informatics and Medical Education, University of Washington, UW Medicine South Lake Union, 850 Republican Street, Building C, Box 358047, Seattle, WA 98109, USA

Phone: 1-206-221-4596

Fax: 1-206-221-2671

Email: luogang@uw.edu

**Acknowledgments**

   We thank Philip J. Brewster, David E. Jones, Michael Conway, and Albert M. Lund for helpful discussions. The work was supported in part by the Natural Science Foundation of China under grant nos. 61463033 and 61262047, and by the Natural Science Foundation of Jiangxi Province under grant no. 20151BAB207028. Most of the work was done during XZ's visit to GL at the University of Utah supported by the China Scholarship Council. The funders had no role in study design, data collection and analysis, decision to publish, or preparation of the manuscript.

**Abstract**

**Purpose**: Machine learning is broadly used for clinical data analysis. Before training a model, a machine learning algorithm must be selected. Also, the values of one or more model parameters termed hyper-parameters must be set. Selecting algorithms and hyper-parameter values requires advanced machine learning knowledge and many labor-intensive manual iterations. To lower the bar to machine learning, miscellaneous automatic selection methods for algorithms and/or hyper-parameter values have been proposed. Existing automatic selection methods are inefficient on large data sets. This poses a challenge for using machine learning in the clinical big data era.

**Methods**: To address the challenge, this paper presents progressive sampling-based Bayesian optimization, an efficient and automatic selection method for both algorithms and hyper-parameter values.

**Results**: We report an implementation of the method. We show that compared to a state of the art automatic selection method, our method can significantly reduce search time, classification error rate, and standard deviation of error rate due to randomization.

**Conclusions**: This is major progress towards enabling fast turnaround in identifying high-quality solutions required by many machine learning-based clinical data analysis tasks.

1. **Introduction**

Machine learning is a key technology for modern clinical data analysis and can be used to support many clinical applications. Examples of clinical machine learning include: (1) Building a model to predict which asthmatic patients are at high risk of hospitalization and enroll them in asthma care management [1, 2]. (2) Building a model to predict appropriate hospital admissions for bronchiolitis patients in the emergency room to guide disposition decisions [3, 4]. (3) Building a classification model to extract symptoms and signs from electronic clinical notes for quality improvement, adverse event surveillance, patient recruitment for clinical trials, and population health metrics [5].

To make machine learning accessible, statistics and computer science researchers have built various open source software tools such as Weka [6], scikit-learn [7], PyBrain [8], RapidMiner, R, and KNIME [9]. These software tools integrate many machine learning algorithms and provide intuitive graphical user interfaces. Despite these efforts, effectively using machine learning is still a difficult task for both computer scientists and layman users with basic computing expertise.

Before training a model for a supervised machine learning problem, the software tool user must perform machine learning model selection by first selecting one out of many applicable algorithms, and then setting the values of one or more model parameters termed hyper-parameters for the selected algorithm. An example of a hyper-parameter is the number of decision trees in a random forest classifier. The algorithm and hyper-parameter values used often impact model error rate by over 40% [10]. The effective ones vary by the machine learning problem and data set [10]. However, selecting algorithms and hyper-parameter values requires advanced machine learning knowledge and many labor-intensive manual iterations, which is non-trivial even for experts in the field of machine learning.

To address this issue, computer science researchers have proposed several dozen automatic selection methods for machine learning algorithms and/or hyper-parameter values. As shown in our review paper [11], most of these methods select either algorithms or hyper-parameter values for a specific algorithm. Only a few published methods [10, 12-14] have been completely implemented and can deal with both many algorithms and an arbitrary number of hyper-parameter value combinations, which is what is most needed in practice. None is efficient on large data sets. The fundamental problem is that a long time is needed for testing a combination of an algorithm and hyper-parameter values on the whole data set. To find an effective combination, these methods test a large number of combinations on the full data set. This can take many hours on a data set with a medium number of attributes and data instances [10]. On a large data set with numerous attributes and/or data instances, these methods' performance will degrade significantly. This poses a challenge for using machine learning in the clinical big data era, as a long search time disallows healthcare researchers to ask a sequence of what-if questions when checking opportunities to adopt machine learning to improve outcomes and cut costs. Due to a lack of access to machine learning experts and the need to build a reasonably accurate machine learning classification model for extracting symptoms and signs from electronic clinical notes, several healthcare researchers in the Veterans Affairs Salt Lake City Health Care System recently approached us to seek help on efficiently and automatically selecting algorithms and hyper-parameter values [5]. To realize personalized medicine, many machine learning problems need to be solved for numerous outcomes and diseases, making search time a bottleneck. Also, outside of healthcare, many machine learning-based data analysis tasks, e.g., in Internet companies, require fast turnaround in identifying high-quality solutions.

To address the challenge, we recently proposed a draft method using sampling to improve the efficiency of automatically selecting both machine learning algorithms and hyper-parameter values [15]. The method will be used to build a tool to enable healthcare researchers to perform machine learning mainly by themselves [15-17]. Our work [15] sketched the method's main ideas, but neither completed the method nor provided any coding implementation. Without further optimization and refinement, the method cannot achieve satisfactory performance. Also, the method's effectiveness is unknown. This paper serves the purpose of filling these gaps.

In this paper, we optimize and complete our efficient and automatic selection method for both machine learning algorithms and hyper-parameter values so that it can achieve better performance. Our method is based on Bayesian optimization [18] and uses progressive sampling [19] as well as several optimizations tailored to automatic machine learning model selection. Hence, we call our method progressive sampling-based Bayesian optimization. We report an implementation of our method. We show that compared to a state of the art automatic selection method, our method can significantly reduce search time, classification error rate, and standard deviation of error rate due to randomization.

This work makes the following contributions:

(1) We present an efficient and automatic method for selecting both machine learning algorithms and hyper-parameter values. As shown in our experimental results, our method can search thousands of combinations of algorithms and hyper-parameter values on a single computer in several hours, regardless of data set size. In comparison, existing automatic selection methods are inefficient on large data sets, where an incomplete search (e.g., of 100 combinations) can take days.

(2) We present several new optimizations tailored to automatic machine learning model selection. With proper extensions, these optimizations can be used to deal with other problems in stochastic optimization.

(3) We present experimental results on many data sets showing that compared to a state of the art automatic selection method, on average our method can reduce search time by 28 fold, classification error rate by 11%, and standard deviation of error rate due to randomization by 36%.

The rest of the paper is organized as follows. In Section 2, we review existing automatic selection methods for both machine learning algorithms and hyper-parameter values, as well as the draft automatic selection method sketched in our previous paper [15]. In Section 3, we describe our main techniques for optimizing and refining the draft method and our complete, progressive sampling-based Bayesian optimization method. In Section 4, we show results from an implementation of our method. We discuss our findings and future work in Section 5 and conclude in Section 6. A roadmap of the paper is shown in Figure 1.

## 2. Background

### 2.1 Review of existing automatic selection methods

In this section, we briefly review existing automatic selection methods for both machine learning algorithms and hyper-parameter values. A detailed review of existing automatic selection methods for algorithms and/or hyper-parameter values is provided in our papers [11, 15].

**2.1.1 Problem statement**

Given a data set $D$, a set of machine learning algorithms $\mathcal{A}$, and a hyper-parameter space $\Lambda$, the goal of machine learning model selection is to identify the algorithm $A^* \in \mathcal{A}$ and hyper-parameter value combination $\lambda^* \in \Lambda$ having the lowest error rate for generalization among all algorithms in $\mathcal{A}$ and hyper-parameter value combinations in $\Lambda$. The generalization performance of an algorithm $A \in \mathcal{A}$ and its hyper-parameter value combination $\lambda \in \Lambda$ is $A_\lambda$'s error rate for new data instances that are not in $D$, where $A_\lambda$ denotes $A$ using $\lambda$. The generalization performance is estimated by $M(A_\lambda, D)$, the error rate attained by $A_\lambda$ when trained and tested on $D$, e.g., via stratified multi-fold cross validation to decrease the possibility of overfitting [6]. Using this estimate, the objective of machine learning model selection is to find $A^*_{\lambda^*} \in \mathrm{argmin}_{A \in \mathcal{A}, \lambda \in \Lambda} M(A_\lambda, D)$.

In reality, because of resource constraints, it is not possible to evaluate every algorithm in $\mathcal{A}$ and hyper-parameter value combination in $\Lambda$ thoroughly and find the combination of an algorithm and hyper-parameter values with the lowest error rate for generalization with complete certainty. Rather, we attempt to identify, within a pre-specified resource limit, a combination of an algorithm and hyper-parameter values that can achieve a low error rate for generalization.

**2.1.2 Hyper-parameters**

A machine learning algorithm can have two types of hyper-parameters: unconditional and conditional. An unconditional hyper-parameter is used at all times. In contrast, a conditional hyper-parameter's relevance relies on another hyper-parameter's value. For example, for support vector machine, the hyper-parameter $\gamma$ is relevant only when the radial basis function kernel is chosen. All hyper-parameters of an algorithm form a directed acyclic graph or tree.

**2.1.3 Bayesian optimization**

Our goal is to automatically select an effective combination of a machine learning algorithm, a feature selection technique (if it is desirable to consider feature selection), and hyper-parameter values. The current way of dealing with this problem [10] is to regard the choice of algorithm and the choice of feature selection technique as new hyper-parameters at the top level and convert this problem to selecting an effective hyper-parameter value combination. Bayesian optimization, a.k.a. sequential model-based optimization [10, 20, 21], is the state of the art method for performing this selection.

Bayesian optimization first builds a regression model $R$ to project a machine learning model's error rate based on hyper-parameter values. $R$ is often a random forest, which can handle both continuous and categorical hyper-parameters and has been shown to perform better than several other ways of making this projection [22]. Bayesian optimization then iterates the following steps: (1) use $R$ to find multiple promising hyper-parameter value combinations to evaluate next; (2) for each such hyper-parameter value combination $\lambda$, train a machine learning model and evaluate its error rate $e$ on the data set at $\lambda$; and (3) update $R$ using the new data points ($\lambda$, $e$). The iteration stops when a pre-chosen stopping criterion is satisfied. In practice, $R$ can be misleading. To ensure robust performance even if this happens, every second hyper-parameter value combination to evaluate next is randomly selected so that new regions of the hyper-parameter space can be examined [10].

As shown in our review paper [11], all published automatic selection methods, which have been completely implemented and can deal with both a wide range of machine learning algorithms and an arbitrary number of hyper-parameter value combinations [10, 12-14], are either based on or are similar to the Auto-WEKA automatic selection method [10]. To obtain a robust estimate of the error rate for generalization of a combination of an algorithm and hyper-parameter values, the method uses up to all 10 folds in 10-fold cross validation to evaluate the combination.

### 2.1.4 Sampling for automatic machine learning model selection

Sampling has been previously used to select either hyper-parameter values for a specific machine learning algorithm [23-29] or algorithms [19, 30-37] (sometimes each with a fixed set of hyper-parameter value combinations [38, 39]), but not for our purpose of selecting hyper-parameter values and algorithms concurrently without limiting the candidate hyper-parameter value combinations for an algorithm to a fixed set. The only exception is the recent work by Li *et al*. [40], which developed an infinitely many armed bandit method using sampling to select both hyper-parameter values and algorithms. This method does not try to avoid search in regions in the hyper-parameter space that contain low-quality combinations. It generally obtains no better search results than the Auto-WEKA automatic selection method [40]. In contrast, our goal is to achieve both better search result quality and search efficiency than the Auto-WEKA automatic selection method. As is typical with Bayesian optimization, our progressive sampling-based Bayesian optimization method tries to avoid search in regions in the hyper-parameter space that contain low-quality combinations.

### 2.2 Our previously sketched automatic selection method

To quickly identify an effective combination of an algorithm and hyper-parameter values for a supervised machine learning problem, we recently designed a draft method to speed up the Bayesian optimization process [15] for selecting both machine learning algorithms and hyper-parameter values. Noticing that evaluating various combinations on a large data set to find an effective one is time-consuming, the method conducts progressive sampling [19], filtering, and fine-tuning to rapidly cut down the search space. Our key idea is to conduct fast trials on limited samples of the data set to drop as many unpromising combinations as quickly as possible, and to give more resources to adjusting promising ones during the Bayesian optimization process. We modify the traditional Bayesian optimization method [10] to consider the factor that change in data sample size affects machine learning model error rate.

As is typical with many search methods, our removal of unpromising combinations is based on heuristics without an absolute guarantee that a high-quality combination would never be dropped. Nevertheless, given a fixed amount of search time, dropping unpromising combinations early enables us to search more combinations, increasing the chance of finding a better final search result. As shown in our empirical evaluation in Section 4, this strategy leads to an overall improvement in both search time and search result quality.

Our method progresses in rounds. In every round, we use both an error rate difference threshold $\tau$ and two non-overlapping, random samples of the data set: the *validation sample* for evaluating each trained model's error rate and the *training sample* for training models. $\tau$ decreases over rounds. As shown in Figure 2, the validation sample remains the same over rounds. The training sample expands over rounds and doubles its size per round. The search space shrinks as the training sample expands (Figure 3). In the last round, we narrow down to a few combinations of machine learning algorithms and hyper-parameter values and select the final combination from them. We build a distinct regression model for every machine learning algorithm and its hyper-parameter values.

*The first round*

In the first round, we use the initial training sample to test various combinations of machine learning algorithms and hyper-parameter values and quickly remove unpromising algorithms. For each algorithm, we test its default hyper-parameter value combination as well as a pre-determined number (e.g., 20) of random hyper-parameter value combinations, if any. For each combination, we use the algorithm, combination, and training sample to train a model and evaluate the model's error rate on the validation sample. The combinations surpassed by the best one at $\geq \tau$ in error rate are regarded as unpromising. Across all combinations tested for the algorithm, the lowest error rate achieved indicates the algorithm's potential. The algorithms surpassed by the best one at $\geq \tau$ in error rate are regarded as unpromising and removed.

*A subsequent round that is not the final one*

In each subsequent round that is not the final one, we decrease the error rate difference threshold $\tau$, expand the training sample, test and adjust the promising combinations of machine learning algorithms and hyper-parameter values on the expanded training sample, and further reduce the search space. We proceed in three steps.

In the first step, for each remaining machine learning algorithm and combination of its hyper-parameter values that appears promising in the prior round, we use the algorithm, combination, and training sample to train a model and evaluate the model's error rate $E_2$ on the validation sample as the combination's error rate estimate for the current round. Due to expansion of the training sample, $E_2$ is usually ≤ the error rate estimate $E_1$ acquired for the algorithm and combination in the prior round. The error rate ratio $r = E_2/E_1$ indicates the degree of decrease in error rate.

In the second step, for each remaining machine learning algorithm, Bayesian optimization is conducted to select and test new hyper-parameter value combinations. We modify the Bayesian optimization method used in Auto-WEKA [10] to consider the factor that expansion of the training sample affects machine learning model error rate. The average error rate ratio *avg_r* is computed for all combinations of the algorithm that appear promising in the previous round. For each combination of the algorithm that looks unpromising in the previous round, we multiply its error rate estimate $E_1$ from the previous round by *avg_r* to obtain its rough error rate estimate for the current round. This saves the overhead of testing the combination on the training and validation samples without significantly affecting which new combinations will be selected by the Bayesian optimization process, as such decisions tend to be impacted mostly by the error rate estimates for the combinations achieving low error rates [22]. Then the regression model is built using all combinations tested for the algorithm so far.

In the third step, we use a method similar to that in the first step to identify unpromising hyper-parameter value combinations and remove unpromising machine learning algorithms.

*The last round*

In the last round, we use the entire data set and best combination of a machine learning algorithm and hyper-parameter values identified to train and evaluate the final model returned by our method.

## 3. Methods

### 3.1 Our main techniques

Our previous paper [15] only sketched the main ideas of our progressive sampling-based Bayesian optimization method for selecting both machine learning algorithms and hyper-parameter values, but did not complete the method. Without further optimization and refinement, the draft method cannot achieve satisfactory performance. Also, the draft method is designed for large data sets and cannot handle small ones. We design eight techniques to optimize and refine the draft method, improve its performance, and make it work for data sets of any size. Figure 4 shows the general structure of our progressive sampling-based Bayesian optimization method. Our complete, progressive sampling-based Bayesian optimization method is presented in Section 3.2. In our implementation, the search process takes five rounds.

In this section, we describe our eight techniques for optimizing and refining the draft method, which are used to better select hyper-parameter value combinations for testing (Technique 1), improve robustness of error rates for generalization estimated for the combinations of machine learning algorithms and hyper-parameter values selected for testing (Technique 2), increase the combinations tested within a limited amount of time (Techniques 3 and 7), avoid testing both invalid and infeasible combinations (Technique 8), help prevent overfitting (Techniques 4 and 6), and avoid testing combinations of feature selection techniques and their hyper-parameter values that cause all or none of the features to be selected (Technique 5). Our techniques and progressive sampling-based Bayesian optimization method use several parameters. We choose their values conservatively and based on our prior experience with other projects, with the goal of keeping the search process efficient without significantly degrading search result quality.

### 3.1.1 Technique 1: Consider distances between combinations in selecting combinations for testing and in obtaining rough error rate estimates for the non-selected ones

Consider a round of the search process that is neither the first nor the last one, and a machine learning algorithm remaining from the previous round. For the hyper-parameter value combinations used in the previous round (i.e., that have been tested so far), our draft automatic selection method chooses some rather than all of them to test on the training and validation samples. The combinations achieving the lowest error rates in the previous round are selected. For the current round, a scaling method is used to obtain rough error rate estimates for the non-selected combinations for reducing computational overhead. The error rate estimates obtained for the selected combinations are more precise than those obtained for the non-selected ones.

The error rates of the non-selected hyper-parameter value combinations need to be estimated precisely enough so that the regression model can guide the search process well. For this purpose, the few hyper-parameter value combinations selected for testing need to cover the hyper-parameter space reasonably well rather than cluster in a small area. To achieve this, we try to select the combinations for testing to be both at least a certain distance away from each other and among the ones with the lowest error rate estimates in the previous round. In comparison, our draft automatic selection method [15] selects the combinations with the lowest error rate estimates in the previous round for testing without imposing any distance constraint.

In our implementation, we use the Hamming distance [41] and choose the default number of hyper-parameter value combinations for testing to be $n_c$=10 and the default distance threshold to be $t_d$=2. We select $n_c$ to be 10 to strike a balance between computational overhead and the need for testing enough combinations. The Hamming distance between two combinations is defined as their number of hyper-parameters with different values. For a numerical hyper-parameter, two values on the original or logarithmic scale are regarded as different if their absolute difference is >1% of the difference between the maximum and minimum values considered for the hyper-parameter. A machine learning algorithm often has only a few hyper-parameters, the number of which specifies the maximum possible Hamming distance between two combinations. Thus, two combinations can be regarded as sufficiently away from each other if the Hamming distance between them is >2.

We select the hyper-parameter value combinations for testing in the following way. If the number of combinations with an error rate <100% that were used in the previous round is ≤$n_c$, all of these combinations are selected for testing. Otherwise, among all of these combinations, we select the $n_c$ combinations for testing in multiple passes. In the first pass, we select the combination with the lowest error rate estimate in the previous round, and then mark every other combination at a distance ≤$t_d$ away from this combination. In each subsequent pass, among all of the combinations not selected or marked, we select the one with the lowest error rate estimate in the previous round, and then mark every other combination at a distance ≤$t_d$ away from this one. We repeat this process until $n_c$ combinations have been selected or all combinations have been selected or marked, whichever occurs first. In the latter case, if the number of selected combinations $n_s$ is <$n_c$, we select from the marked combinations the top $n_c$–$n_s$ ones with the lowest error rate estimates in the previous round so that $n_c$ combinations are chosen in total.

Let $\lambda_i$ (1≤$i$≤$g$) denote the hyper-parameter value combinations selected for testing, $g$ denote the number of them, and $r_i$ denote the error rate ratio of $\lambda_i$. For each non-selected combination with an error rate estimate of 100% from the previous round, we keep its error rate estimate at 100% for the current round. For each other non-selected combination $\lambda_u$, we multiply its error rate estimate from the previous round by a factor $r_u$ to obtain its rough error rate estimate for the current round. If the rough error rate estimate is >100%, it is set to 100%. Using inverse distance weighting that helps improve estimation accuracy, $r_u$ is computed as a weighted average of the selected combinations' error rate ratios:

$r_u = \sum_{i=1}^{g} w_{i,u} \times r_i / \sum_{i=1}^{g} w_{i,u}$.

Here, the weight $w_{i,u}$ for $\lambda_i$ is set to $1/distance(\lambda_u, \lambda_i)$ [42], reflecting the intuition that the further away $\lambda_i$ is from $\lambda_u$, the less weight $\lambda_i$ should have. In case distance($\lambda_u$, $\lambda_j$)=0 for a specific $j$ (1≤$j$≤$g$), we set $r_u$=$r_j$. In comparison, our draft automatic selection method [15] uses an ordinary average rather than a weighted average to compute $r_u$.

An error rate ratio $r_i$ (1≤$i$≤$g$) can have an extreme value, e.g., when the hyper-parameter value combination $\lambda_i$ has its test terminated and its error rate estimate set to 100% in the current round (see Section 3.1.3). To prevent the factor $r_u$ from being overly influenced by an extreme $r_i$, we force each $r_i$ to fall into the range [0.25, 2.5]. If an $r_i$ is <0.25, we set it to 0.25. If an $r_i$ is >2.5, we set it to 2.5. In each round that is neither the first nor the last one, the training sample doubles in size [15], which often impacts a combination's error rate estimate by no more than two times. The lower and upper bounds of the range are set conservatively reflecting this insight.

### 3.1.2 Technique 2: On a small data set, use multiple folds to estimate each selected combination's error rate for generalization

In any given round of the search process that is not the final one, our draft automatic selection method uses a single pair of training and validation samples to estimate the error rate for generalization of every combination of a machine learning algorithm and hyper-parameter values selected for testing. The obtained estimates are used to identify unpromising combinations and to guide the search direction. When the data set is small, the validation sample is small. Due to insufficient testing of the trained models, the obtained estimates can be non-robust. This affects the final search result's quality.

To address this problem, we use a technique similar to multi-fold cross validation and differentiate between two cases: a small data set and a large data set. On a small data set, a single validation sample cannot ensure robustness of the error rates for generalization estimated for the combinations of machine learning algorithms and hyper-parameter values selected for testing. Instead, multiple folds are used to increase the obtained estimates' robustness. Due to the data set's small size, doing this will not incur excessive computational overhead. On a large data set, a single validation sample is relatively large and can ensure that the obtained estimates are relatively robust. A single fold is used to reduce computational overhead.

Let $n$ denote the number of data instances in the data set. We use $m$=min(5,000, $n$) randomly sampled data instances of the data set to conduct training and validation in the first four rounds of the search process (i.e., all rounds except for the final one). We impose an upper bound of 5,000 on $m$ to control the search process' computational overhead, while making the upper bound large enough for the training and validation samples to sufficiently represent the entire data set.

*The case with a small data set*

When the data set is small, we use a pre-determined integer $k>1$ and perform $k$-fold progressive sampling. The search process' computational overhead increases with $k$. To avoid excessive computational overhead, $k$ should not be too large. To ensure robustness of the error rates for generalization estimated for the combinations of machine learning algorithms and hyper-parameter values selected for testing, $k$ should not be too small. In our implementation, we choose $k$'s default value to be 3 to strike a balance between computational overhead and robustness of the estimated error rates for generalization.

We perform $k$-fold progressive sampling in the following way. We randomly divide the $m$ data instances into $k$ parts of roughly the same size. In the case of a classification problem, for every class of the dependent variable, each part contains roughly the same number of instances of the class.

In any given round of the search process that is not the final one, we use $k$ folds. As shown in Figure 5, in the $i$-th ($1 \leq i \leq k$) fold, the $i$-th part of the $m$ data instances forms the validation sample. The union of all other parts of the $m$ data instances forms the largest training set. In the first four rounds (i.e., all rounds except for the final one), the training sample doubles its size per round [19] and includes 12.5%, 25%, 50%, and 100% of data instances in the largest training set, respectively. For each combination of a machine learning algorithm and hyper-parameter values selected for testing, we use the training sample and combination to train a model and estimate the model's error rate on the validation sample. The average value of the estimated model error rate across all $k$ folds serves as the combination's estimated error rate in this round.

*The case with a large data set*

When the data set is large, we perform 1-fold progressive sampling. 1/3 of the $m$ data instances form the validation sample. The other 2/3 of the $m$ data instances form the largest training set. Other details mirror the case of a small data set.

*Determining whether the data set is small or large*

A machine learning algorithm's running time usually increases at least linearly with the number of features (a.k.a. independent variables) and superlinearly with the number of data instances. Hence, the product of these two numbers of the data set roughly reflects the search process' computational overhead when $k$-fold progressive sampling is used. To avoid excessive computational overhead of the search process, we can switch from $k$-fold to 1-fold progressive sampling when this product is large. Based on this insight, we use this product to differentiate between small and large data sets. When this product is larger than $10^6$, the data set is regarded as large. Otherwise, the data set is regarded as small. A data set with 5,000 data instances (the upper bound of $m$) and 200 features has a product of $10^6$.

### 3.1.3 Technique 3: In each round of the search process, limit the time that can be spent on testing a combination

In practice, there are usually several effective combinations of machine learning algorithms, feature selection techniques, and hyper-parameter values achieving similar error rates. Some can be tested much faster than others. Finding any of them would suffice our purpose. To ensure that the search process can test many combinations within reasonable time without being unduly delayed on a single combination, we limit the time that can be spent on testing a combination. This can improve the final search result's quality. Alternatively, this can expedite the search process.

In each round of the search process, for each test of a combination of a machine learning algorithm, a feature selection technique, and hyper-parameter values on a fold, we impose a limit $L_f$ on feature selection time and a limit $L_t$ on model training time. Like Auto-WEKA [10], once the time spent on feature selection exceeds $L_f$, the test is terminated and the model's estimated error rate is set to 100%. We store the feature selection technique and its hyper-parameter values in a cache shared by all algorithms. In case feature selection is completed within $L_f$, once the time spent on model training exceeds $L_t$, model training is stopped and the partially trained model's error rate is estimated on the validation sample.

Before testing a combination of a machine learning algorithm, a feature selection technique, and hyper-parameter values $\lambda$, we first check whether the feature selection technique and its hyper-parameter values are in the cache. A match is found when the Hamming distance is zero. In this case, we know that the time spent on feature selection in the test will most likely exceed $L_f$ even without testing $\lambda$. We set $\lambda$'s estimated error rate for generalization to 100%, skip testing $\lambda$, and use ($\lambda$, 100%) as a new data point to update the regression model.

Each feature selection technique uses a feature search method and a feature evaluator [10]. To expedite search in the cache, we build an index on the categorical hyper-parameter specifying the choice of feature evaluator as well as an index on each numerical hyper-parameter of a feature search method.

In our implementation, for a small data set, we start with $L_f$=10 seconds and $L_t$=10 seconds in the first round of the search process. For a large data set, we start with $L_f$=20 seconds and $L_t$=20 seconds in the first round. These thresholds are chosen to strike a balance between computational overhead and the goal of not terminating too many tests. As the training sample expands and the time required for testing a combination of a machine learning algorithm, a feature selection technique, and hyper-parameter values increases over rounds, we increase both $L_f$ and $L_t$ by 50% per round (Figure 6). We do not double both $L_f$ and $L_t$ per round, as otherwise in later rounds, both $L_f$ and $L_t$ will become so large that overly time-consuming tests cannot be quickly stopped. As the combinations selected for testing change over rounds, this technique keeps pushing the search direction towards combinations that can both achieve low error rates and be tested relatively quickly. If a combination has its test terminated and its error rate estimate set to 100%, its error rate ratio will be large. Then the inverse distance weighting mechanism will cause its neighboring combinations not selected for testing to have high rough estimates of the error rate for the current round. Consequently, subsequent searches will be prompted to avoid this neighborhood.

The Auto-WEKA automatic selection method [10] limits the time that can be spent on testing a combination of a machine learning algorithm, a feature selection technique, and hyper-parameter values on the whole data set: 15 minutes for feature selection and 150 minutes for model training. Since a long time is needed for conducting a test on a large data set, the Auto-WEKA automatic selection method uses long timeouts and cannot quickly stop overly time-consuming tests. In contrast, in the initial few rounds of the search process, our technique uses short timeouts and small samples of the data set to quickly identify combinations that cannot be tested fast enough. Then in later rounds, we can avoid spending a lot of time testing these combinations on large samples of the data set.

In the Bayesian optimization process, Snoek *et al.* [21] proposed optimizing the ratio of expected improvement in error rate to the time required for testing a hyper-parameter value combination on the whole data set. This also serves the purpose of continuously pushing the search direction towards combinations that can both achieve low error rates and be tested relatively quickly.

### 3.1.4 Technique 4: Perform cross validation to select the final combination

In our draft automatic selection method, the same validation sample with a moderate number of data instances is used throughout the search process, where many tests are conducted with a potential for overfitting [43-46]. To help prevent overfitting, in the final round of the search process, we add a step of performing $h$-fold cross validation to evaluate the best machine learning algorithms and hyper-parameter value combinations found so far. In our implementation, we set $h=10$ for a small data set. To avoid excessive computational overhead, we set $h=3$ rather than 10 for a large data set.

Ideally, the cross validation should be performed on a separate sample of the data set that has never been used before. This is possible only if the data set has enough data instances. Recall that 5,000 is the maximum number of data instances that can used for conducting training and validation in the previous rounds (Section 3.1.2). On a data set with ≤5,000 data instances, we perform cross validation on the whole data set. On a data set with >5,000 data instances, we perform cross validation on a random sample of 5,000 data instances of the data set. These 5,000 data instances are first chosen from the ones that have never been used before. If not enough data instances are available there, the remainder of the 5,000 data instances is chosen from the ones used before. In the case of a classification problem, for every class of the dependent variable, we ensure that the proportion of instances of it in the 5,000 data instances is roughly the same as that in the whole data set.

For each combination of a machine learning algorithm and hyper-parameter values, *h*-fold cross validation will produce *h* error rate estimates, one per fold. A single extreme value in these *h* error rate estimates, if any, can greatly affect their average value. To address this issue, when comparing two combinations, a pairwise approach rather than the traditional average value approach is used to obtain more stable result. We compare the two comparisons' error rate estimates in each fold. The combination winning in more folds wins the overall comparison.

For each machine learning algorithm remaining from the previous round, we choose the top 10 hyper-parameter value combinations with the lowest error rate estimates in the previous round, if any. The number 10 is selected to strike a balance between computational overhead and the goal of not missing high-quality combinations. Each pair of an algorithm and a top combination is compared with each other pair. The pair winning the largest number of times is the final one chosen by our automatic selection method. If two or more champions exist, their average values of the *h* error rate estimates in the *h* folds are used to break ties. If two or more champions still exist after this step, their error rate estimates from the previous round are used to break ties. Finally if two or more champions still exist, training and validating time across the *h* folds is used to break ties. The pair with the shortest training and validating time is chosen by our automatic selection method.

### 3.1.5 Technique 5: Inspect results upon completing feature selection

Some combinations of feature selection techniques and their hyper-parameter values will cause all or none of the features to be selected. Using all features is the same as conducting no feature selection. When no feature is selected, the machine learning software such as Weka will use the ZeroR algorithm [6], which predicts the majority class (for a classification problem) or the average value (for a regression problem), regardless of which machine learning algorithm is specified by the automatic selection process.

Consider a test of a combination of a machine learning algorithm that is not ZeroR, a feature selection technique, and hyper-parameter values $\lambda$. If all or none of the features are selected, we can use the algorithm to train a model and evaluate the model's error rate *e*. However, using ($\lambda$, *e*) as a new data point to update the regression model will mislead the subsequent search process. If no feature is selected, *e* reflects ZeroR's error rate rather than the algorithm's error rate. If all features are selected, the updated regression model based on ($\lambda$, *e*) can suggest testing another combination of the same feature selection technique and hyper-parameter values of this technique, along with a different algorithm and its hyper-parameter values. In this case, conducting feature selection is unnecessary. Instead, we should skip feature selection and directly start model training.

To avoid misleading the subsequent search process, we inspect results upon completing feature selection. If all or none of the features are selected for a combination of a machine learning algorithm, a feature selection technique, and hyper-parameter values $\lambda$, we set $\lambda$'s estimated error rate for generalization to 100%, skip model training, and use ($\lambda$, 100%) as a new data point to update the regression model. The data point ($\lambda$, 100%) hints to avoid future testing of the combination of the same feature selection technique and hyper-parameter values of this technique. This helps save time. For the same purpose, we store the feature selection technique and its hyper-parameter values in a cache shared by all algorithms.

Before testing a combination of a machine learning algorithm, a feature selection technique, and hyper-parameter values $\lambda$, we first check whether the feature selection technique and its hyper-parameter values are in the cache. A match is found when the Hamming distance is zero. In this case, we know that most likely all or none of the features will be selected even without testing $\lambda$. We set $\lambda$'s estimated error rate for generalization to 100%, skip testing $\lambda$, and use ($\lambda$, 100%) as a new data point to update the regression model. To expedite search in the cache, we build indexes on certain hyper-parameters as described in Section 3.1.3.

In a given round of the search process, suppose for a machine learning algorithm we want to select $q$ combinations of feature selection techniques and hyper-parameter values for testing. The combinations causing all or none of the features to be selected do not count towards the goal of reaching $q$ combinations. In the first round, the selection process is repeated until we have made 200 trials or found $q$ combinations, none of which causes all or none of the features to be selected, whichever occurs first. A combination hit by the cache does not count towards the goal of making 200 trials. The threshold 200 is conservatively chosen for making enough trials to meet the goal of reaching $q$ combinations without incurring excessive computational overhead. In each other round, the maximum number of allowed trials is changed from 200 to $q+5$. We allow more trials in the first round than in any other round, because in the first round the training sample is small and conducting a trial is fast.

The cache is shared by all algorithms. In the first round as algorithms get processed, more and more combinations of feature selection techniques and their hyper-parameter values accumulate in the cache. At the end of the first round, for each pair of a machine learning algorithm and a combination of a feature selection technique and its hyper-parameter values $\lambda_1$ in the cache, we select a random hyper-parameter value combination $\lambda_2$ of the algorithm and use (($\lambda_1$, $\lambda_2$), 100%) as a new data point to update the regression model for the algorithm. This helps the regression model to guide subsequent searches away from regions in the hyper-parameter space where all or none of the features will be selected. A similar approach is used to handle the combinations of feature selection techniques and their hyper-parameter values in the cache described in Section 3.1.3.

### 3.1.6 Technique 6: Penalize the use of feature selection

A data analytic pipeline contains one or more parts. Feature selection can be one of these parts. The pipeline's complexity grows with the number of parts. In particular, using feature selection is more complex than not using it. The more complex the pipeline, the more likely overfitting will occur. To help prevent overfitting, we impose a penalty on the use of feature selection as a form of regularization. For a combination of a machine learning algorithm, a feature selection technique, and hyper-parameter values, we multiply its estimated error rate by a pre-determined constant >1 for updating the regression model if feature selection is used in the combination. In our implementation, we choose the constant's default value to be 1.1 to give enough of a penalty to, but not over-penalize, the use of feature selection.

### 3.1.7 Technique 7: Penalize the use of a meta or ensemble algorithm

Machine learning algorithms fall into three categories: base, ensemble, and meta [10]. A base algorithm such as support vector machine can be used by itself. A meta algorithm such as AdaBoost takes as an input a base algorithm and its hyper-parameter values. An ensemble algorithm such as stacking takes multiple base algorithms as input. Meta and ensemble algorithms run more slowly than base algorithms. To expedite the search process, we impose a penalty on the use of a meta or ensemble algorithm as a form of regularization. For a meta or ensemble algorithm, the number of base algorithms used in it $n_b$ reflects its complexity. For a combination of a meta or ensemble algorithm and hyper-parameter values, we multiply its estimated error rate by $1+0.02\times n_b$ for updating the regression model. The number 0.02 is chosen to give enough of a penalty to, but not over-penalize, the use of a meta or ensemble algorithm.

### 3.1.8 Technique 8: Avoid testing both invalid and infeasible combinations

Before starting the search process, we already know that certain combinations of machine learning algorithms and hyper-parameter values are invalid or infeasible to test. Here are some examples:

(1) Certain hyper-parameter values conflict with each other and cannot be used together. For example, the naive Bayes classification algorithm cannot adopt kernel density estimator and supervised discretization, both of which are used for handling continuous features, at the same time. As another example, each feature selection technique uses a feature search method and a feature evaluator [10]. The ranker feature search method ranks individual features. It is not compatible with any feature subset evaluator [6].

(2) In the presence of a large number of features, certain hyper-parameter value combinations become too time-consuming to test. For example, principal component analysis becomes overly time-consuming to conduct when there are more than 2,000 features.

(3) When the dependent variable is categorical with many classes, certain hyper-parameter value combinations become too time-consuming to test. For instance, using exhaustive error-correction codes is one way to transform a multi-class classification problem into multiple two-class ones. In the presence of many (e.g., >10) classes, using exhaustive codes becomes infeasible because its computational complexity grows exponentially with the number of classes [6].

It is a poor use of time to test an invalid or infeasible combination of a machine learning algorithm and hyper-parameter values, which usually leads to a runtime exception, an out of memory error, or a timeout error. To avoid such testing, we manually mark both invalid and infeasible combinations and store their information in the form of rules in the automatic machine learning model selection software. In the search process, before testing a combination, we first check whether it is invalid or infeasible. If so, we skip testing it.

**3.2 The complete, progressive sampling-based Bayesian optimization method**

In this section, we present the details of our complete, progressive sampling-based Bayesian optimization method. Our method proceeds in rounds.

**3.2.1 The first round**

In the first round, we test every applicable machine learning algorithm. For each algorithm, we test its default hyper-parameter value combination as well as a pre-determined number of random hyper-parameter value combinations, if any. In our implementation, we choose this number's default value to be 20.

To quickly reduce the search space, a large percentage of machine learning algorithms needs to be removed. For this purpose, we first regard the algorithms surpassed by the best one at $\geq \tau$ in error rate as unpromising and remove them, where $\tau$=0.5 is the error rate difference threshold. Then if >40% of all algorithms remain, we continue to remove additional algorithms and keep only the top 40% of all algorithms with the lowest error rates. The numbers 0.5 and 40% are chosen to strike a balance between removing enough algorithms in the first round for search efficiency and the goal of not removing high-quality algorithms too early. In any case, we try to keep at least three algorithms by not removing more algorithms so that <min($n_a$, 3) algorithms remain, where $n_a$ is the number of all algorithms. The search process can be performed relatively quickly with three algorithms. There is no need to drop the number of algorithms below three in the first round. This may happen otherwise, e.g., if $n_a$ is small.

**3.2.2 A subsequent round that is not the final one**

In each subsequent round that is not the final one, we decrease the error rate difference threshold $\tau$ by multiplying it by 0.8 (Figure 7), expand the training sample, increase both the limit $L_f$ on feature selection time and the limit $L_t$ on model training time, test and adjust the promising combinations of machine learning algorithms and hyper-parameter values on the expanded training sample, and further reduce the search space. The number 0.8 is chosen to strike a balance between removing enough algorithms in the current round for search efficiency and the goal of not removing high-quality algorithms too early.

For each machine learning algorithm remaining from the previous round, we proceed in three steps. In the first step, we use the technique in Section 3.1.1 to select several hyper-parameter value combinations of the algorithm used in the previous round for testing and obtain their error rate estimates for the current round. In the second step, for each combination of the algorithm used in the previous round but not selected for testing in the first step, we use the technique in Section 3.1.1 to obtain its rough error rate estimate for the current round. In the third step, we build a regression model using all combinations tested for the algorithm so far. Then we conduct $C$ cycles of Bayesian optimization. We start with $C=3$ in the second round and decrease $C$ by one per round (Figure 8), considering the factor that a test's computational overhead increases over rounds. In each cycle of Bayesian optimization, like Auto-WEKA [10, 20], 10 new combinations are selected for testing and used to revise the regression model. To examine new regions of the hyper-parameter space, every other combination is selected at random. A combination selected for testing in the third step can be one used in the previous round but not selected for testing in the first step of the current round.

Finally, we use a method similar to that in the first round to identify and remove unpromising machine learning algorithms. The main difference here is that the targeted percentage of algorithms that will be kept is changed from 40% to 70%. Many low-quality algorithms have already been removed in the previous rounds. It is more difficult to identify unpromising algorithms now. Thus, compared to that in the first round, we become more conservative in removing algorithms to avoid dropping high-quality ones too early.

**3.2.3 The final round**

In the final round, we use the technique in Section 3.1.4 to select the final combination of a machine learning algorithm and hyper-parameter values. We use this combination to build a model on the whole data set as the final model returned by our automatic selection method.

**3.2.4 Additional details on handling various types of algorithms**

In each round of the search process, we test base, ensemble, and meta algorithms. Base algorithms are tested before ensemble and meta algorithms. The base algorithms removed in the first round will not be used by any ensemble or meta algorithm. In contrast, the base algorithms removed in any subsequent round can still be used by an ensemble or meta algorithm in future rounds. In the first round, the decision of which base algorithms should be removed is made before we start to test ensemble and meta algorithms. The decision of which ensemble and meta algorithms should be removed is made at the end of the round. In any subsequent round that is not the final one, the remaining base, ensemble, and meta algorithms are pooled together. The decision of which algorithms should be removed is made at the end of the round for all pooled algorithms simultaneously.

Among all machine learning algorithms, random forest and support vector machine are frequently among the top performing ones in terms of error rate [47]. To prevent these two algorithms from being dropped too early without sufficient testing of their potential, we do not remove them in the first two rounds of the search process.

For the machine learning algorithm of support vector machine, our draft automatic selection method regards the use of each type of kernel as a separate algorithm [15]. We discovered in our testing that this technique provided no benefit and do not use it in our implementation.

**4. Results**

We did an implementation of our progressive sampling-based Bayesian optimization method by modifying the source code of Auto-WEKA [10]. In this section, we present results from our implementation.

**4.1 Experiment description**

As shown in our review paper [11], all published automatic selection methods, which have been completely implemented and can handle both a wide range of machine learning algorithms and an arbitrary number of hyper-parameter value combinations [10, 12-14], are either based on or are similar to the Auto-WEKA automatic selection method [10]. Auto-WEKA handles more machine learning algorithms than hyperopt-sklearn [13]. In our experiments, we compared our progressive sampling-based Bayesian optimization method with the Auto-WEKA automatic selection method. As an initial study, our purpose was to demonstrate the effectiveness and feasibility of using progressive sampling and its related optimizations to select both algorithms and hyper-parameter values. Hence, we did not test any additional technique that can improve both our method and the Auto-WEKA automatic selection method in similar ways. Two examples of such techniques are: (1) using distributed computing to expedite the search process on a computer cluster [48] and (2) using knowledge from experiments with prior machine learning problems to select a good starting point for Bayesian optimization [12, 49, 50]. We leave it for future work to integrate such techniques and other optimizations into our method.

Our experiments used basically the same setup as that in the Auto-WEKA paper [10]. We considered all 39 classification algorithms and feature selection techniques in the standard Weka package [6]. We used 21 well-known, non-clinical benchmark data sets (Table 1) adopted in the Auto-WEKA paper [10] and available at [51]. Table 2 shows samples from one of these data sets. In addition, we included six well-known, clinical benchmark data sets from the University of California, Irvine machine learning repository [52]: Arcene, Arrhythmia, Cardiotocography, Diabetic Retinopathy Debrecen, Mammographic Mass, and Parkinson Speech. Each data set has a training/test split. The training data were used in the search process. The test data were used to evaluate the error rate of the final model returned by the automatic selection method. For the Auto-WEKA method, each run on a data set had a total time budget of 30 hours.

Our measurements were performed on a Dell PowerEdge R430 server (210-ADLO) with two 12-core 2.5GHz processors, 256GB main memory, and running the CentOS Linux 6.7 operating system.

**4.2 Overall results of the search process**

For each data set, we ran both the Auto-WEKA automatic selection method and our progressive sampling-based Bayesian optimization method five times with different random seeds. Every parameter used in our method was set to its default value. For each automatic selection method and data set, three metrics are presented in Table 3: time spent on the search process, number of distinct combinations of machine learning algorithms and hyper-parameter values tested, and the final model's error rate on the test data. For each metric, its average value and standard deviation over the five runs are shown as average value ± standard deviation.

*Time spent on the search process*

The Auto-WEKA automatic selection method always consumed its total time budget of 30 hours. In comparison, the search process of our progressive sampling-based Bayesian optimization method took much less time: on average 44 times faster on small data sets and 5 times faster on large data sets. Our method's search process tends to take longer on large data sets than on small ones. The speedup gained by our method over the Auto-WEKA method in the search process varies from 3 on the MNIST Basic data set to 334 on the Parkinson Speech data set.

*Number of distinct combinations tested*

Due to its use of the sampling technique, our progressive sampling-based Bayesian optimization method can test a combination of a machine learning algorithm and hyper-parameter values much faster than the Auto-WEKA automatic selection method. This is particularly the case with large data sets. Compared to the Auto-WEKA method, our method tested many more distinct combinations in the search process. On small data sets, on average our method tested 3 times more distinct combinations than the Auto-WEKA method. On 8 of the 16 small data sets (Mammographic Mass, Car, Yeast, German Credit, Diabetic Retinopathy Debrecen, Parkinson Speech, Cardiotocography, and KR-vs-KP), the Auto-WEKA method tested more distinct combinations than our method by spending more time on the search process. On large data sets, on average our method tested 37 times more distinct combinations than the Auto-WEKA method. The increase in the number of distinct combinations tested that our method gained over the Auto-WEKA method varies from 12 times on the KDD09-Appentency data set to 128 times on the ROT. MNIST+BI data set. Due to its inefficiency, the Auto-WEKA method tested much fewer combinations than what would be needed for exploring the search space sufficiently to even a moderate extent.

*The final model's error rate on the test data*

On almost all of the data sets, our progressive sampling-based Bayesian optimization method achieved a lower error rate on the test data than the Auto-WEKA automatic selection method. There were five exceptions. On the data sets Car, Parkinson Speech, and KDD09-Appentency, our method achieved the same error rate as the Auto-WEKA method. On the small data sets Yeast, Abalone, and Wine Quality, the Auto-WEKA method achieved a lower error rate than our method, but the difference was small (<1%).

On small data sets, on average our progressive sampling-based Bayesian optimization method achieved a 4% lower error rate than the Auto-WEKA automatic selection method. On large data sets, on average our method achieved a 21% lower error rate than the Auto-WEKA method. On the MNIST Basic data set, our method reached the maximum reduction in error rate: by 64%. Our method has a more significant advantage on large data sets than on small ones. This is unsurprising. Since a long time is required to test a combination of a machine learning algorithm and hyper-parameter values on a large data set, the Auto-WEKA method tested inadequate combinations and thus produced low-quality final search results on large data sets.

*Standard deviation of error rate due to randomization*

On small data sets, both the Auto-WEKA automatic selection method and our progressive sampling-based Bayesian optimization method tested many distinct combinations of machine learning algorithms and hyper-parameter values. On average our method achieved a 16% lower standard deviation of error rate due to randomization than the Auto-WEKA method. On large data sets, on average our method achieved a 65% lower standard deviation of error rate due to randomization than the Auto-WEKA method, because our method tested many more distinct combinations than the Auto-WEKA method. The reduction in standard deviation of error rate due to randomization that our method gained over the Auto-WEKA method varies from 0% on the Car, Parkinson Speech, and KDD09-Appentency data sets to 97% on the ROT. MNIST+BI data set.

*Feature selection results of the final combination*

Table 4 presents feature selection results of the final combinations of machine learning algorithms and hyper-parameter values produced by the automatic selection methods. In the final combinations produced by the Auto-WEKA automatic selection method, feature selection was conducted in 96 of the 135 runs (71%). 86% of the time (83 of 96), feature selection was conducted unnecessarily, causing all or none of the features to be selected in 79 and 4 runs, respectively. In comparison, in the final combinations produced by our progressive sampling-based Bayesian optimization method, feature selection was conducted in only 7 of the 135 runs (5.2%) and never caused all or none of the features to be selected.

There are two possible reasons why feature selection was rarely conducted in the final combinations of machine learning algorithms and hyper-parameter values produced by our progressive sampling-based Bayesian optimization method. First, to help prevent overfitting, we impose a penalty on the use of feature selection as a form of regularization. Second, several top-performing algorithms, such as support vector machine and random forest, have an internal feature selection mechanism embedded within them. If our method selects any of them as the final algorithm, a separate feature selection part in the data analytic pipeline often becomes unnecessary.

On the 27 data sets on average, compared to the Auto-WEKA automatic selection method, our progressive sampling-based Bayesian optimization method reduced search time by 28 fold, error rate by 11%, and standard deviation of error rate due to randomization by 36%. Our method gained similar performance advantages on both clinical and non-clinical data sets. This is major progress towards enabling fast turnaround in identifying high-quality solutions required by many machine learning-based data analysis tasks.

**4.3 Impacts of individual techniques**

Our progressive sampling-based Bayesian optimization method uses the following main techniques:

(1) **Technique 1**: Consider distances between hyper-parameter value combinations in selecting combinations for testing and in obtaining rough error rate estimates for the non-selected ones.

(2) **Technique 2**: On a small data set, use multiple folds to estimate the error rate for generalization of each selected combination of a machine learning algorithm and hyper-parameter values.

(3) **Technique 3**: In each round of the search process, limit the time that can be spent on testing a combination.

(4) **Technique 4**: Perform cross validation to select the final combination.

(5) **Technique 5**: Inspect results upon completing feature selection.

(6) **Technique 6**: Penalize the use of feature selection.

(7) **Technique 7**: Penalize the use of a meta or ensemble algorithm.

(8) **Technique 8**: Avoid testing both invalid and infeasible combinations.

In this section, we describe the impacts of individual techniques on search result quality and search time. We conducted a set of experiments. In each experiment, we removed one technique while keeping the others constant. For each data set, we conducted each experiment five times with different random seeds and computed each metric's average value over the five runs. To remain consistent with our experimental methodology described in Section 4.1, each run on a data set had a total time budget of 30 hours. In the description below, we explicitly point out every case in which the search process could not finish in 30 hours.

On each of the four small data sets Car, KR-vs-KP, Parkinson Speech, and Shuttle, the final model's error rate on the test data was <0.5% when all techniques were used. If any technique or sub-technique was not used, the error rate would either remain the same or increase. In case of increase, the increase was typically by several times due to the small baseline error rate. This large degree of increase was much larger than the degree of error rate change on any of the other 23 data sets and would dominate the average degree of error rate change on all 27 data sets. To show the impacts of individual techniques more clearly, in the following description, we omit the four small data sets Car, KR-vs-KP, Parkinson Speech, and Shuttle and focus on the other 23 data sets. Table 5 summarizes the impacts of individual techniques.

*Impacts of Technique 1*

When Technique 1 was not used, the hyper-parameter value combinations with the lowest error rate estimates in the previous round of the search process were selected for testing. All of them were equally weighted in obtaining rough error rate estimates for the non-selected combinations. As explained in Section 3.1.3, the inverse distance weighting mechanism helps the search process avoid regions in the hyper-parameter space in which the combinations are overly time-consuming to test. Without this mechanism, the search process is more likely to run into such regions, increasing search time. When Technique 1 was not used and compared to the case that all techniques were used, on all data sets on average, search time increased by 2%. The final model's error rate on the test data remained the same.

*Impacts of the sub-techniques in Technique 1*

Technique 1 uses two sub-techniques: (a) trying to select the hyper-parameter value combinations for testing to be both at least a certain distance away from each other and among the ones with the lowest error rate estimates in the previous round of the search process, and (b) using inverse distance weighting to obtain rough error rate estimates for the combinations not selected for testing. We conducted two experiments to show the impacts of these two sub-techniques. In either experiment, we removed one sub-technique while keeping the other constant.

When the first sub-technique was not used, the hyper-parameter value combinations with the lowest error rate estimates in the previous round were selected for testing. These combinations could possibly cluster in a small area with a poor coverage of the hyper-parameter space, creating the potential of the error rates of the non-selected combinations being poorly estimated. This reduced the regression model's capability to guide the right search direction and impacted the final search result's quality. When the first sub-technique was not used and compared to the case that all techniques were used, on all data sets on average, the final model's error rate on the test data increased by 2%.

When the second sub-technique was not used, all hyper-parameter value combinations selected for testing were equally weighted in obtaining rough error rate estimates for the non-selected combinations. As explained in Section 3.1.3, the inverse distance weighting mechanism helps the search process avoid regions in the hyper-parameter space in which the combinations are overly time-consuming to test. Without this mechanism, the search process is more likely to run into such regions, increasing search time. This also decreases the thoroughness of searching other regions that contain high-quality and fast-to-test combinations, impacting the final search result's quality. When the second sub-technique was not used and compared to the case that all techniques were used, on all data sets on average, search time increased by 1%. The final model's error rate on the test data increased by 3%.

*Impacts of Technique 2*

When Technique 2 was not used, 1-fold progressive sampling was performed regardless of data set size. This impacts both search time and search result quality on small data sets. As the number of folds decreases, the computational overhead decreases. The error rates for generalization estimated for the combinations of machine learning algorithms and hyper-parameter values become less robust, affecting the final search result's quality. When Technique 2 was not used and compared to the case that all techniques were used, on small data sets on average, search time was reduced by 41%. The final model's error rate on the test data increased by 3.5%.

*Impacts of Technique 3*

When Technique 3 was not used, in any given round of the search process, the same limit used in the Auto-WEKA automatic selection method [10] was imposed on the time that could be spent on testing a combination of a machine learning algorithm and hyper-parameter values. Each run on a data set had a total time budget of 30 hours. In this case, many overly time-consuming tests could not be quickly stopped, increasing search time. On 18 of the 23 data sets (78%) including all the large ones, the search process could not finish in 30 hours. On the other five small data sets on average, compared to the case that all techniques were used, search time increased by 1397%. The final model's error rate on the test data remained the same.

*Impacts of Technique 4*

When Technique 4 was not used, in the final round of the search process, the combination of a machine learning algorithm and hyper-parameter values with the lowest error rate estimate in the previous round was selected as the final combination. Cross validation time was saved at the expense of increased likelihood of overfitting. When Technique 4 was not used and compared to the case that all techniques were used, on all data sets on average, search time was reduced by 15%. The final model's error rate on the test data increased by 4%.

*Impacts of the sub-technique in Technique 4*

Technique 4 uses a sub-technique: using a pairwise approach to compare two combinations of machine learning algorithms and hyper-parameter values. We conducted an experiment to show the impacts of this sub-technique. When this sub-technique was not used, the traditional average value approach was used to compare two combinations, creating the potential of the comparison result being overly influenced by a single extreme value of the error rate estimated for a combination.

Compared to the case that all techniques were used, on all data sets on average, the final model's error rate on the test data increased by 1%.

*Impacts of Technique 5*

When Technique 5 was not used, no result was inspected upon completing feature selection. Many combinations of feature selection techniques and their hyper-parameter values that caused none of the features to be selected were tested and used to update the regression model, misleading the subsequent search process. Also, in a given round of the search process, for each machine learning algorithm, the combinations of feature selection techniques and hyper-parameter values that caused all or none of the features to be selected counted towards the goal of reaching the targeted number of combinations selected for testing. Thus, fewer combinations were likely to be tested, decreasing search time. When Technique 5 was not used and compared to the case that all techniques were used, on all data sets on average, search time was reduced by 12%. The final model's error rate on the test data increased by 2%.

*Impacts of the sub-techniques in Technique 5*

Technique 5 uses two sub-techniques: (a) when all or none of the features are selected for a combination of a machine learning algorithm, a feature selection technique, and hyper-parameter values, storing the feature selection technique and its hyper-parameter values in a cache shared by all algorithms; and (b) at the end of the first round, using the information in the cache to form additional data points to update the regression model for every algorithm. We conducted two experiments to show the impacts of these two sub-techniques.

When the first sub-technique was not used, the feature selection techniques and their hyper-parameter values that caused all or none of the features to be selected were not stored in a cache. Then in each round of the search process that was not the final one, we could not use the cache to identify such feature selection techniques and their hyper-parameter values. For each machine learning algorithm, the combinations of feature selection techniques and hyper-parameter values that caused all or none of the features to be selected did not count towards the goal of reaching the targeted number of combinations selected for testing. Hence, more combinations were likely to be tested, increasing search time. When the first sub-technique was not used and compared to the case that all techniques were used, on all data sets on average, search time increased by 2%. The final model's error rate on the test data increased by 2%.

When the second sub-technique was not used, at the end of the first round, no additional data point based on the information in the cache was used to update the regression model for any machine learning algorithm. This reduced the regression model's capability of guiding subsequent searches away from regions in the hyper-parameter space where all or none of the features would be selected, and increased the likelihood that an insufficient number of effective trials were made in some of the rounds. When the second sub-technique was not used and compared to the case that all techniques were used, on all data sets on average, search time increased by 1%. The final model's error rate on the test data increased by 2%.

*Impacts of Technique 6*

When Technique 6 was not used, the use of feature selection was not penalized. Consequently, feature selection was more likely to be used, which increased the likelihood of overfitting. When Technique 6 was not used and compared to the case that all techniques were used, on all data sets on average, the final model's error rate on the test data increased by 1%.

*Impacts of Technique 7*

When Technique 7 was not used, the use of a meta or ensemble algorithm was not penalized. Consequently, meta and ensemble algorithms were more likely to be tested in the search process. Meta and ensemble algorithms run more slowly than base algorithms and can result in overfitting. When Technique 7 was not used and compared to the case that all techniques were used, on all data sets on average, search time increased by 6%. The final model's error rate on the test data increased by 1%.

*Impacts of Technique 8*

When Technique 8 was not used, both invalid and infeasible combinations of machine learning algorithms and hyper-parameter values were tested, resulting in a poor use of time. On the data sets Abalone, Arcene, and Amazon, the search process crashed because of out of memory errors. Compared to the case that all techniques were used, on the other 20 data sets on average, search time increased by 2%. The final model's error rate on the test data increased by 2%.

In summary, all eight techniques in our progressive sampling-based Bayesian optimization method are necessary. If any of them is not used, the error rate and/or search time will increase. No single technique impacts the error rate by more than 4%. For our progressive sampling-based Bayesian optimization method to significantly reduce the error rate, all eight techniques need to be used together.

**5. Discussion**

Automatically selecting machine learning algorithms and hyper-parameter values is a challenging problem. Many existing attempts to tackle this problem take the analytical approach of developing mathematical models to better guide the search direction [11]. Using this approach alone is unlikely to make automatic machine learning model selection fast enough for regular, practical use on clinical big data. Instead of going through the analytical route, in this work we take the system approach of improving search efficiency by using techniques like sampling and caching. Sampling of the data set produces imprecise error rate estimates of combinations of algorithms and hyper-parameter values. At the same time, sampling significantly increases the number of combinations that can be tested in a fixed amount of time. The benefit gained from more tests outweighs the loss of precision in estimating error rates of combinations, the impact of which can be minimized via careful use of various techniques. As a result, sampling, if properly used, can greatly improve both search speed and the final search result's quality. The key to speeding up the search process without sacrificing the final search result's quality is to correctly remove as many unpromising items as quickly as possible. Example items include algorithms, hyper-parameter value combinations, features, and feature selection techniques. This insight provides general guidance for us to design additional techniques to further improve search efficiency in the future.

Our test results in Section 4 show major performance advantages of our system approach for automating machine learning model selection. Recently, the healthcare researchers in the Veterans Affairs Salt Lake City Health Care System used our automatic machine learning model selection method successfully for a clinical research project, which required building a reasonably accurate machine learning classification model for extracting symptoms and signs from electronic clinical notes [5]. This gives another piece of evidence for our system approach's effectiveness.

Future studies will investigate other techniques to further improve search efficiency without degrading search result quality. Example techniques include: (1) removing unpromising features early, (2) dropping unpromising feature selection techniques early, (3) delaying time-consuming tests to the end of the search process [16, 17], (4) performing lower-precision, such as 32-bit rather than 64-bit, operations in model training and evaluation to expedite the first few rounds of the search process, (5) being more conservative in the first few rounds of the search process in removing algorithms, such as deep learning, that often require much training data to show their potential, and (6) using special hardware to expedite some key processing steps.

Our current method mainly uses heuristic techniques. In the future, it would be interesting to investigate developing theoretical underpinnings for improving our method. For example, a theory can be useful for determining the initial training sample size in the first round of the search process and how to expand the training sample over rounds in an optimal way for Bayesian optimization. As another example, in an intermediate round, a theory can be useful for determining which hyper-parameter value combinations should be selected for testing and how to obtain rough error rate estimates for the non-selected ones. Compared to using any single approach alone, we would expect combining analytical, software system, and hardware system approaches will give us the largest gains in both search speed and the final search result's quality.

**6. Conclusions**

We optimized and completed an efficient and automatic selection method for both machine learning algorithms and hyper-parameter values. Our experiments show that our method greatly improves both search efficiency and search result quality compared to a state of the art automatic selection method.

**Competing interests**

The authors declare that they have no competing interests.

**Authors' contributions**

XZ participated in designing the study, wrote the paper's first draft, did the computer coding implementation, and performed experiments. GL conceptualized and designed the study, conducted literature review, and rewrote the whole paper. Both authors read and approved the final manuscript.

**Acknowledgments**

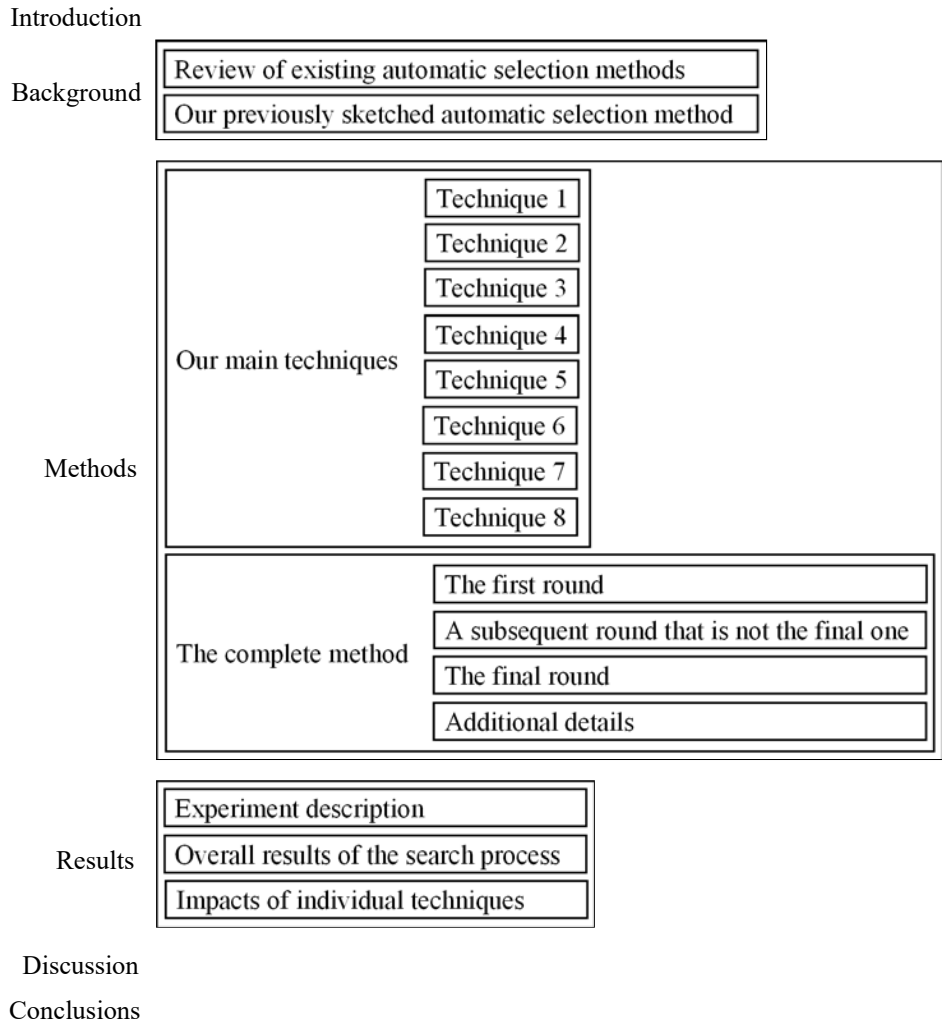

**Fig. 1** A roadmap of this paper

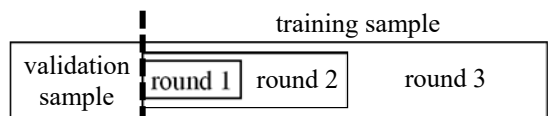

**Fig. 2** An illustration of progressive sampling used in our automatic machine learning model selection method

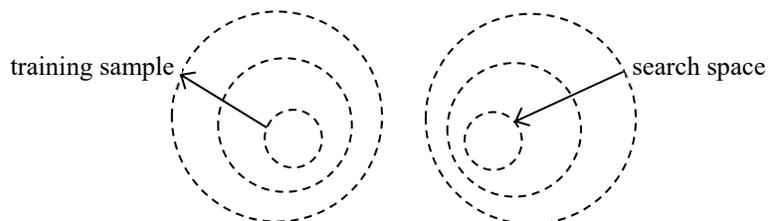

**Fig. 3** The relationship between the training sample and search space in our progressive sampling-based Bayesian optimization method

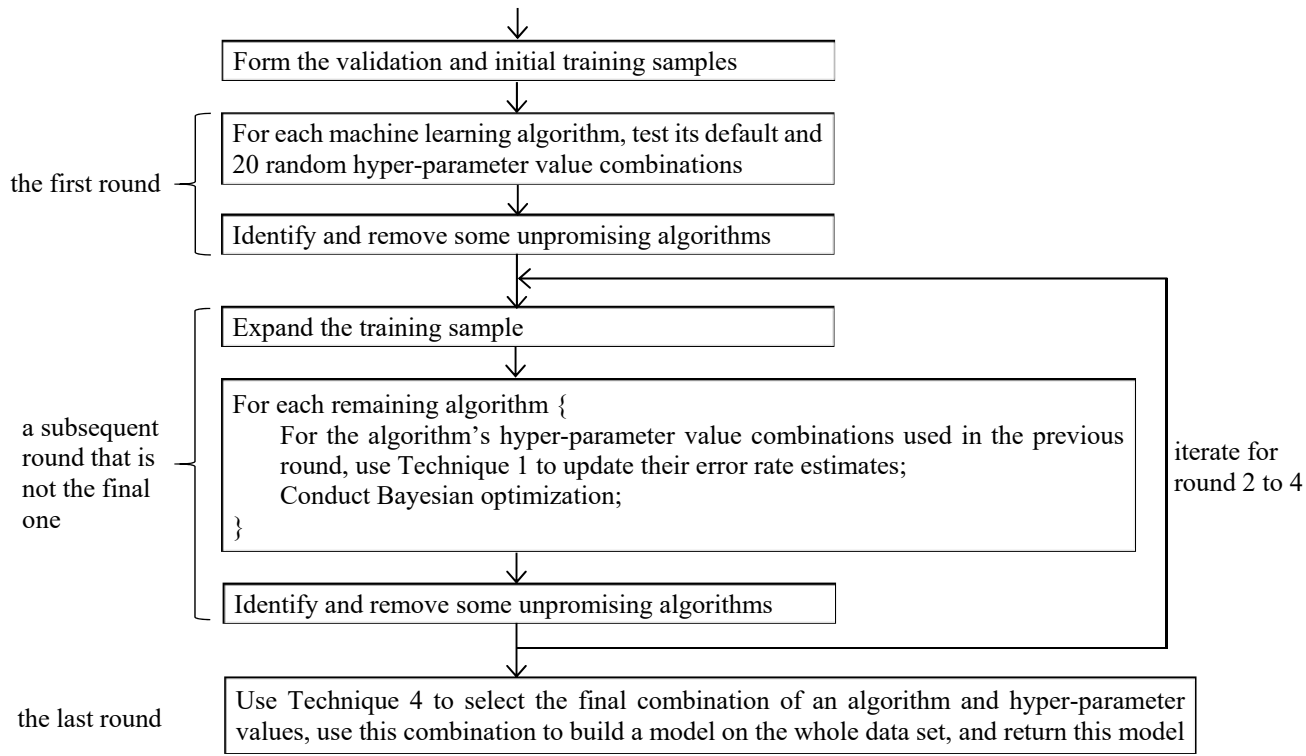

**Fig. 4** The algorithm framework of our progressive sampling-based Bayesian optimization method

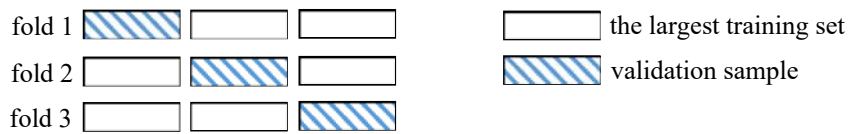

**Fig. 5** In the first four rounds of the search process, the training and validation set partitioning for 3-fold progressive sampling on a small data set

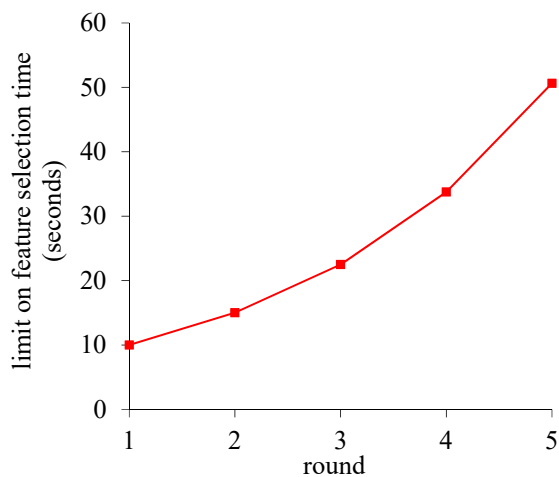

**Fig. 6** On a small data set, the limit on feature selection time over rounds

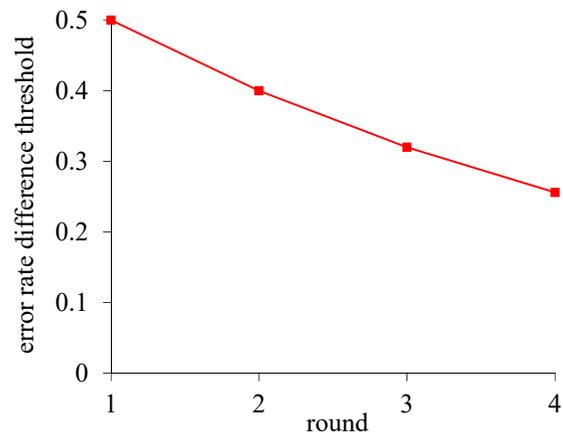

**Fig. 7** The error rate difference threshold over rounds

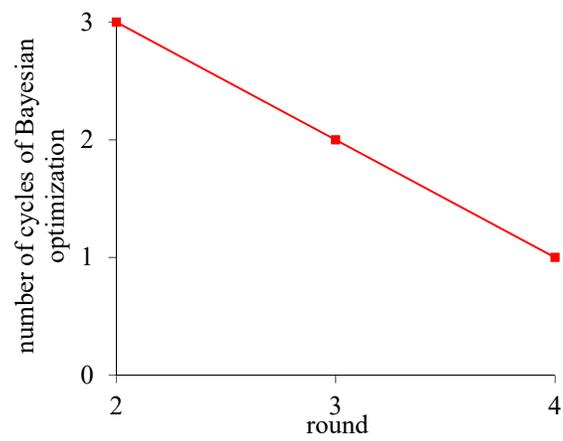

**Fig. 8** The number of cycles of Bayesian optimization over rounds

Table 1. The data sets used (ordered by # of training instances × # of attributes). Small data sets are shown in the first 16 rows. Large data sets are shown in the last 11 rows.

| Name | # of training instances | # of attributes | # of test instances | # of classes |
|---|---:|---:|---:|---:|
| Mammographic Mass | 673 | 6 | 288 | 2 |
| Car | 1,209 | 6 | 519 | 4 |
| Yeast | 1,038 | 8 | 446 | 10 |
| German Credit | 700 | 20 | 300 | 2 |
| Diabetic Retinopathy Debrecen | 806 | 20 | 345 | 2 |
| Parkinson Speech | 728 | 26 | 312 | 2 |
| Abalone | 2,923 | 8 | 1,254 | 28 |
| Cardiotocography | 1,488 | 23 | 638 | 3 |
| Wine Quality | 3,425 | 11 | 1,469 | 11 |
| KR-vs-KP | 2,237 | 37 | 959 | 2 |
| Arrhythmia | 316 | 279 | 136 | 16 |
| Waveform | 3,500 | 40 | 1,500 | 3 |
| Semeion | 1,115 | 256 | 478 | 10 |
| Shuttle | 43,500 | 9 | 14,500 | 7 |
| Secom | 1,096 | 591 | 471 | 2 |
| Madelon | 1,820 | 500 | 780 | 2 |
| Arcene | 100 | 10,000 | 100 | 2 |
| Convex | 8,000 | 784 | 50,000 | 2 |
| KDD09-Appentency | 35,000 | 230 | 15,000 | 2 |
| Dexter | 420 | 20,000 | 180 | 2 |
| MNIST Basic | 12,000 | 784 | 50,000 | 10 |
| ROT. MNIST+BI | 12,000 | 784 | 50,000 | 10 |
| Amazon | 1,050 | 10,000 | 450 | 49 |
| Gisette | 4,900 | 5,000 | 2,100 | 2 |
| CIFAR-10-Small | 10,000 | 3,072 | 10,000 | 10 |
| Dorothea | 805 | 100,000 | 345 | 2 |
| CIFAR-10 | 50,000 | 3,072 | 10,000 | 10 |

Table 2. Samples from the Car data set.

| buying | maint | doors | persons | lug_boot | safety | class |
|---|---|---|---|---|---|---|
| vhigh | vhigh | 2 | 2 | small | high | unacc |
| high | low | 4 | 4 | med | high | acc |
| low | low | 2 | more | med | high | good |

Table 3. Overall results of the search process.

| Data set | Time spent (hours) | | # of distinct combinations tested | | Error rate on the test data (%) | |
|---|---|---|---|---|---|---|
| | Auto-WEKA | Our method | Auto-WEKA | Our method | Auto-WEKA | Our method |
| Mammographic Mass | 30 | 0.3±0.3 | 4,085±1,798 | 1,615±25 | 17.36±0.49 | 16.94±0.38 |
| Car | 30 | 2.0±0.4 | 1,954±1,050 | 1,524±35 | 0.00±0.00 | 0.00±0.00 |
| Yeast | 30 | 1.7±0.4 | 6,675±2,721 | 1,602±23 | 38.29±1.51 | 38.88±1.20 |
| German Credit | 30 | 1.0±0.3 | 3,364±952 | 1,602±29 | 28.40±1.26 | 27.00±1.11 |
| Diabetic Retinopathy Debrecen | 30 | 1.4±0.3 | 4,621±1,434 | 1,745±39 | 25.57±0.66 | 24.58±1.58 |
| Parkinson Speech | 30 | 0.09±0.01 | 3,570±250 | 1,669±35 | 0.00±0.00 | 0.00±0.00 |
| Abalone | 30 | 4.3±0.5 | 675±181 | 1,501±20 | 72.99±0.30 | 73.25±0.47 |
| Cardiotocography | 30 | 0.5±0.2 | 3,578±1,962 | 1,600±34 | 1.63±0.29 | 1.54±0.13 |
| Wine Quality | 30 | 4.2±0.8 | 1,429±823 | 1,633±33 | 33.97±0.75 | 34.12±0.70 |
| KR-vs-KP | 30 | 1.5±0.6 | 2,192±1,522 | 1,574±23 | 0.42±0.24 | 0.40±0.11 |
| Arrhythmia | 30 | 3.83±0.59 | 259±72 | 1,464±39 | 33.24±1.83 | 30.59±1.12 |
| Waveform | 30 | 3.5±0.6 | 801±248 | 1,596±21 | 14.33±0.15 | 14.29±0.12 |
| Semeion | 30 | 4.7±0.7 | 408±103 | 1,594±9 | 5.58±0.82 | 5.03±0.21 |
| Shuttle | 30 | 1.7±0.5 | 275±92 | 1,610±22 | 0.010±0.004 | 0.008±0.003 |
| Secom | 30 | 1.1±0.1 | 119±90 | 1,699±47 | 8.13±0.57 | 7.83±0.09 |
| Madelon | 30 | 2.6±0.3 | 529±261 | 1,763±49 | 22.08±2.52 | 18.31±1.47 |
| Arcene | 30 | 3.37±0.68 | 108±37 | 1,635±10 | 18.00±3.16 | 16.40±1.52 |
| Convex | 30 | 5.5±0.4 | 72±49 | 1,766±73 | 26.24±4.93 | 23.37±0.87 |
| KDD09-Appentency | 30 | 6.0±0.5 | 155±127 | 1,783±72 | 1.74±0.00 | 1.74±0.00 |
| Dexter | 30 | 4.4±0.6 | 98±14 | 1,569±28 | 8.33±1.24 | 5.11±1.20 |
| MNIST Basic | 30 | 11.1±0.7 | 53±16 | 1,674±23 | 10.14±4.40 | 3.70±0.46 |
| ROT. MNIST+BI | 30 | 7.0±0.3 | 13±13 | 1,637±17 | 63.31±5.76 | 55.81±0.16 |
| Amazon | 30 | 8.9±0.8 | 53±19 | 1,561±41 | 38.31±14.73 | 24.31±1.91 |
| Gisette | 30 | 9.4±0.3 | 43±25 | 1,586±31 | 2.31±0.55 | 1.89±0.29 |
| CIFAR-10-Small | 30 | 10.0±0.5 | 46±21 | 1,546±34 | 67.87±5.36 | 57.42±0.52 |
| Dorothea | 30 | 7.3±0.9 | 52±45 | 1,625±25 | 6.29±2.09 | 5.74±0.48 |
| CIFAR-10 | 30 | 10.0±0.9 | 29±24 | 1,529±72 | 62.02±7.68 | 52.83±0.78 |

Table 4. Feature selection (FS) results of the final combinations.

| Data set | Auto-WEKA | | | Our method |
|---|---|---|---|---|
| | # of times FS was conducted | # of times FS selected all features | # of times FS selected no feature | # of times FS was conducted |
| Mammographic Mass | 4 | 4 | 0 | 0 |
| Car | 4 | 4 | 0 | 0 |
| Yeast | 2 | 2 | 0 | 0 |
| German Credit | 3 | 3 | 0 | 0 |
| Diabetic Retinopathy Debrecen | 3 | 3 | 0 | 0 |
| Parkinson Speech | 4 | 3 | 0 | 1 |
| Abalone | 4 | 4 | 0 | 0 |
| Cardiotocography | 4 | 3 | 0 | 0 |
| Wine Quality | 4 | 4 | 0 | 0 |
| KR-vs-KP | 2 | 2 | 0 | 0 |
| Arrhythmia | 3 | 1 | 0 | 0 |
| Waveform | 4 | 4 | 0 | 0 |
| Semeion | 5 | 5 | 0 | 0 |
| Shuttle | 3 | 3 | 0 | 0 |
| Secom | 4 | 1 | 0 | 0 |
| Madelon | 4 | 1 | 0 | 4 |
| Arcene | 3 | 3 | 0 | 0 |
| Convex | 4 | 4 | 0 | 0 |
| KDD09-Appentency | 5 | 0 | 3 | 0 |
| Dexter | 4 | 4 | 0 | 0 |
| MNIST Basic | 3 | 2 | 0 | 0 |
| ROT. MNIST+BI | 2 | 2 | 0 | 0 |
| Amazon | 4 | 4 | 0 | 1 |
| Gisette | 4 | 4 | 0 | 0 |
| CIFAR-10-Small | 3 | 3 | 0 | 0 |
| Dorothea | 4 | 3 | 1 | 1 |
| CIFAR-10 | 3 | 3 | 0 | 0 |
| Total | 96 | 79 | 4 | 7 |

Table 5. The impacts of individual techniques on search time and the final model's error rate on the test data. "-" means no change.

| Technique | Impact of not using the technique | |
|---|---|---|
| | search time | error rate |
| Technique 1 | ↑2% | - |
| Technique 2 | ↓41% | ↑3.5% |
| Technique 3 | ↑1397% | - |
| Technique 4 | ↓15% | ↑4% |
| Technique 5 | ↓12% | ↑2% |
| Technique 6 | - | ↑1% |
| Technique 7 | ↑6% | ↑1% |
| Technique 8 | ↑2% | ↑2% |